\documentclass[10pt, a4paper]{article}

\usepackage[final]{lrec2026} 

\usepackage{adjustbox}
\usepackage{booktabs}
\usepackage{caption}

\title{When Words Don’t Mean What They Say: Figurative Understanding in Bengali Idioms}

\name{
\\
    \textbf{Adib Sakhawat}$^{1}$
\quad
\textbf{Shamim Ara Parveen}$^{2}$
\quad
\textbf{Md Ruhul Amin}$^{2}$
\\
\textbf{Shamim Al Mahmud}$^{2}$
\quad
\textbf{Md Saiful Islam}$^{2}$
\quad
\textbf{Tahera Khatun}$^{2}$\\ 
\\
$^{1}$Islamic University of Technology\quad $^{2}$National University of Bangladesh
 \\
    {\tt adibsakhawat@iut-dhaka.edu}
}

\address{}
\abstract{
Figurative language understanding remains a significant challenge for Large Language Models (LLMs), especially for low-resource languages. To address this, we introduce a new idiom dataset, a large-scale, culturally-grounded corpus of 10,361 Bengali idioms. Each idiom is annotated under a comprehensive 19-field schema, established and refined through a deliberative expert consensus process, that captures its semantic, syntactic, cultural, and religious dimensions, providing a rich, structured resource for computational linguistics. To establish a robust benchmark for Bangla figurative language understanding, we evaluate 30 state-of-the-art multilingual and instruction-tuned LLMs on the task of inferring figurative meaning. Our results reveal a critical performance gap, with no model surpassing 50\% accuracy, a stark contrast to significantly higher human performance (83.4\%). This underscores the limitations of existing models in cross-linguistic and cultural reasoning. By releasing the new idiom dataset and benchmark, we provide foundational infrastructure for advancing figurative language understanding and cultural grounding in LLMs for Bengali and other low-resource languages.
\\ \newline
}

\begin{document}

\maketitleabstract

\section{Introduction}

The computational modeling of figurative language presents a significant and persistent challenge in Natural Language Processing (NLP). As noted in recent surveys, research on idioms has proceeded along two parallel tracks psycholinguistics and computational linguistics with little crossover \cite{flor-etal-2025-survey}. While high-resource languages like English have benefited from extensive corpus development for both type-based idiom detection (e.g., MAGPIE \cite{haagsma-etal-2020-magpie}) and token-based contextual disambiguation (e.g., IDIX \cite{sporleder-etal-2010-idioms}), this progress remains starkly uneven. The difficulty of idiom annotation, even in high-resource settings \cite{street-etal-2010-like}, is compounded for the over 1.5 billion speakers of South Asian languages, which remain critically under-resourced in NLP \cite{arora-etal-2022-computational}.

Bangla (Bengali), the world's seventh most spoken language, epitomizes this resource scarcity. Prior work on Bangla figurative language has been limited in scope and focus. Early efforts centered on rule-based translation systems for small sets of idioms \cite{khatun-et-al-2020} or statistical identification of general multi-word expressions (MWEs) \cite{chakraborty2014identifyingbengalimultiwordexpressions}. More recently, datasets for figurative language have emerged, such as Alankaar \cite{rakshit-flanigan:2025:RANLP1}, which provides a valuable resource of 2,500 examples for metaphor identification. However, a large-scale, computationally-oriented \textit{idiom} corpus with deep cultural annotation has been unavailable. Existing resources are either smaller in scale, focus on different linguistic phenomena like metaphors, or lack the cultural grounding necessary to interpret idioms in context, a difficulty magnified by Bangla's complex morphology \cite{deo2006typological}.

\begin{figure}[!h]
    \centering
    \includegraphics[width=.75\linewidth]{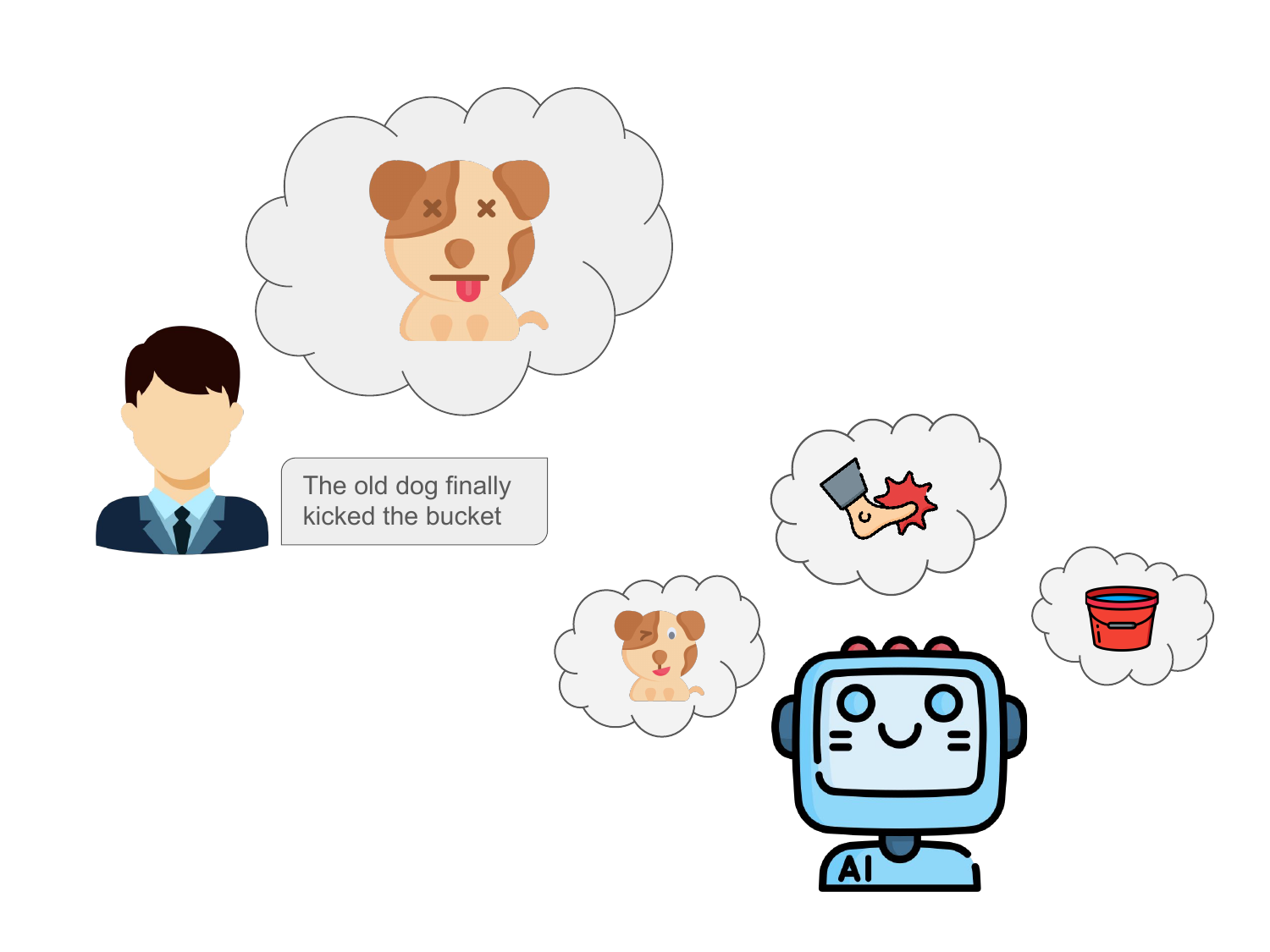}
    \caption{Conceptual illustration contrasting human and AI interpretations of idioms (‘kicked the bucket’).}
    \label{fig:full_process}
\end{figure}

While existing multilingual resources offer some utility, they possess notable limitations. Datasets like LIdioms \cite{moussallem-etal-2018-lidioms} and IMIL \cite{agrawal-etal-2018-beating} focus primarily on facilitating cross-lingual transfer rather than providing the monolingual depth and fine-grained cultural annotation necessary for a nuanced understanding of idioms in their native context. Similarly, commendable regional efforts such as the Konidioms corpus for Konkani \cite{shaikh-etal-2024-konidioms} operate at a scale insufficient for developing and evaluating robust computational models.

To address this critical gap, we introduce a new idiom dataset, a large-scale, culturally-grounded corpus of 10,361 Bangla idioms. This work offers four principal contributions:

\begin{enumerate}
    \item \textbf{Scale and Coverage:} With 10,361 meticulously annotated entries, the new idiom dataset is the largest and most comprehensive figurative language resource for Bangla.
    
    \item \textbf{Rich Annotation Schema:} We introduce a novel 19-field annotation framework. It includes literal and figurative meanings (Bangla/English), usage examples, and deep cultural metadata (historical origins, religious context, socio-spatial relevance).
    
    \item \textbf{Extensive Benchmarking:} We establish the first comprehensive benchmark for Bangla figurative language understanding. We evaluated 30 state-of-the-art LLMs on a subset of \textbf{100 idioms hand-picked by expert consensus}. For a human baseline, \textbf{15 Bangladeshi high school (6) and undergrad (9) students} took the test in a \textbf{controlled environment}, achieving a \textbf{mean score of 83.4\%}. The results show a stark performance chasm: while humans averaged 83.4\% accuracy, no LLM surpassed the 50\% mark, exposing a critical deficiency in current models.
    
    \item \textbf{Cultural Grounding:} The new idiom dataset systematically integrates cultural context as a primary annotation layer, enabling novel research at the intersection of computational linguistics and cultural analytics.
\end{enumerate}

The new idiom dataset and our accompanying benchmark analysis provide foundational infrastructure to advance a new generation of culturally-aware NLP applications for Bangla, including machine translation, sentiment analysis, and educational tools. By making this resource publicly available, we aim to catalyze targeted research efforts to bridge the significant performance gap in multilingual figurative language understanding.\footnote{https://www.kaggle.com/datasets/sakhadib/bangla-bagdhara}

\section{Related Work}

Computational modeling of idiomaticity presents significant challenges, particularly for resource-constrained languages with substantial linguistic diversity. While English-centric approaches dominate the field, recent multilingual expansions remain inadequate for languages such as Bangla. A survey of existing literature reveals critical gaps in scale, annotation depth, and cultural metadata that our work aims to address \cite{flor-etal-2025-survey}.

The \textbf{MAGPIE corpus} \cite{haagsma-etal-2020-magpie} established a benchmark with over 50,000 English expressions, demonstrating the feasibility of large-scale idiom annotation. The \textbf{FLUTE dataset} \cite{chakrabarty-etal-2022-flute} introduced a paradigm shift through textual explanations across 9,000 figurative instances using human-AI collaboration. Similarly, the \textbf{IDIX corpus} provided context-aware annotations for English idioms, which facilitated token-level disambiguation of literal versus figurative usage \cite{sporleder-etal-2010-idioms}. Work by \cite{street-etal-2010-like} on the American National Corpus further highlighted the subjective nature of idiom identification and the importance of including varied syntactic structures, such as prepositional phrases.

Multilingual efforts have achieved limited success. The \textbf{LIdioms dataset} \cite{moussallem-etal-2018-lidioms} provided cross-lingual linkages for 815 idioms across five languages, while the \textbf{IMIL corpus} \cite{agrawal-etal-2018-beating} expanded to 2,200 idioms across eight Indian languages. However, these resources remain limited in both scale and annotation depth compared to their English counterparts.

\begin{table}[h]
\centering
\begin{adjustbox}{max width=\columnwidth}
\begin{tabular}{lcccc}
\hline
\textbf{Corpus} & \textbf{Size} & \textbf{Languages} & \textbf{Annotation Fields} \\
\hline
MAGPIE & 50,000+ & 1 (English) & 5 \\
FLUTE & 9,000 & 1 (English) & 8 \\
LIdioms & 815 & 5 & 6 \\
IMIL & 2,200 & 8 & 7 \\
Konidioms & 1,597 & 1 (Konkani) & 9 \\
Alankaar & 2,500 & 1 (Bangla) & 2 \\
Our Dataset & \textbf{10,361} & \textbf{1 (Bangla)} & \textbf{19} \\
\hline
\end{tabular}
\end{adjustbox}
\caption{Comparative Analysis of Major Idiom Corpora}
\label{tab:corpus_comparison}
\end{table}

Significant regional disparities persist, with South Asian languages severely underrepresented despite their substantial speaker populations. While Konkani received attention through the \textbf{Konidioms corpus} \cite{shaikh-etal-2024-konidioms} with 1,597 entries, Bengali serving over 300 million speakers has lacked dedicated computational resources. Early work on Bengali multi-word expressions (MWEs) by \cite{chakraborty2014identifyingbengalimultiwordexpressions} highlighted their non-compositional nature, and a rule-based translation system for Bengali idioms was proposed by \cite{khatun-et-al-2020}. More recently, the \textbf{Alankaar} dataset provided 2,500 annotated examples of metaphors in Bangla, exploring the cultural challenges of translating figurative language \cite{rakshit-flanigan:2025:RANLP1}. However, a large-scale, comprehensively annotated idiom-specific resource has been missing. This scarcity reflects broader challenges in South Asian NLP \cite{Baker2003ConstructingCO}, where Bangla's linguistic complexity \cite{deo2006typological} and semantic non-compositionality intersect with cultural contexts \cite{arora-etal-2022-computational}.

Annotation methodologies have evolved from basic frameworks like \textbf{IDIX} \cite{sporleder-etal-2010-idioms} and \textbf{PIE} \cite{haagsma-etal-2020-magpie} to sophisticated multimodal approaches with \textbf{V-FLUTE} \cite{chakrabarty-etal-2022-flute}. However, contemporary datasets often focus on narrow semantic distinctions, with FLUTE's explanations constrained to natural language inference tasks and LIdioms prioritizing cross-linguistic connectivity over semantic depth.

Cultural metadata remains critically underdeveloped. While some datasets include basic domain classifications, systematic annotation of historical significance, religious context, and socio-spatial dimensions is largely absent. Cross-linguistic approaches frequently prioritize direct translation over cultural adaptation \cite{flor-etal-2025-survey}, neglecting essential cultural models. Recent research emphasizes culturally informed annotation schemas \cite{Pakray_Gelbukh_Bandyopadhyay_2025} for morphologically complex languages.

Both open-source and proprietary large language models significantly struggle with understanding figurative language \cite{khoshtab-etal-2025-comparative}, particularly for low-resource languages due to training data gaps. Open-source models demonstrate particularly pronounced limitations compared to some proprietary systems \cite{khoshtab-etal-2025-comparative}.

The development of the new idiom dataset addresses these gaps for Bangla through a large-scale, comprehensively annotated idiom resource with a 19-field schema, establishing new standards for low-resource language resource development.

\section{Methodology}

The construction of the new idiom dataset and its subsequent evaluation followed a systematic, multi-phase protocol designed to ensure linguistic authenticity, computational robustness, and scholarly integrity. Figure~\ref{fig:full_process} shows the complete pipeline of the new idiom dataset construction, annotation, and evaluation process.

\subsection{Dataset Architecture and Schema}

The dataset's core is a comprehensive 19-field JSON schema, engineered to capture the multidimensional nature of idiomatic expressions beyond semantics. Each of the 10,361 entries adheres to this structure, which is organized into four key logical groups:

\begin{enumerate}
    \item \textbf{Semantic and Cross-Lingual Mapping:} Fields such as \texttt{idiom}, \texttt{literal\_meaning}, \texttt{figurative\_meaning\_bn/en}, and \texttt{similar\_in\_english} establish a triad mapping denotation to connotation, providing cross-lingual anchors for translation and analogical reasoning.
    \item \textbf{Pragmatic Contextualization:} \texttt{example\_sentences\_in\_bangla/english}, \texttt{usage\_domain}, and \texttt{tags} situate idioms in authentic linguistic environments, which is essential for training context-aware models.
    \item \textbf{Sociocultural and Affective Metadata:} Boolean flags for \texttt{historical}, \texttt{religious}, and \texttt{cultural\_significance}, alongside \texttt{sentiment} polarity, encode vital symbolic and affective context.
    \item \textbf{Diachronic and Ethnolinguistic Information:} The \texttt{scape} field defines the socio-spatial domain (e.g., urban, rural), while \texttt{history} and \texttt{note} preserve provenance and expert ethnolinguistic commentary.
\end{enumerate}

\begin{figure}[!htb]
    \centering
    \includegraphics[width=.75\linewidth]{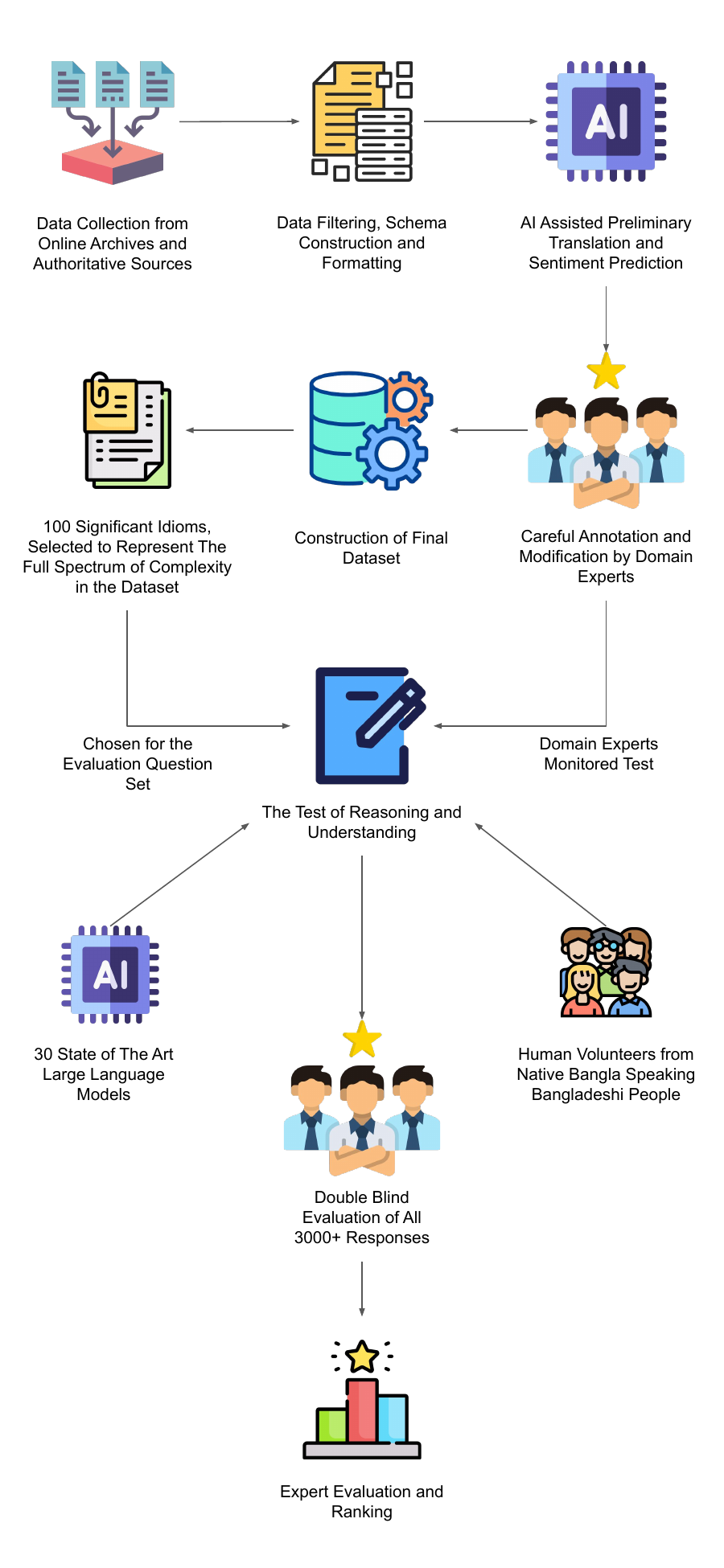}
    \caption{The complete pipeline of the new idiom dataset construction, annotation, and evaluation process.}
    \label{fig:full_process}
\end{figure}

The complete schema is detailed in Table~\ref{tab:dataset_schema}.

\begin{table}[h!]
\centering
\begin{adjustbox}{max width=\columnwidth}
\begin{tabular}{@{}lll@{}}
\toprule
\textbf{Field} & \textbf{Type} & \textbf{Language} \\
\midrule
id & integer & English \\
idiom & string & Bangla \\
alternative\_idioms & array & Bangla \\
literal\_meaning & string & Bangla \\
figurative\_meaning\_bn & string & Bangla \\
figurative\_meaning\_en & string & English \\
similar\_in\_english & array & English \\
example\_sentences\_in\_bangla & array & Bangla \\
example\_sentences\_in\_english & array & English \\
usage\_domain & array & Mixed \\
tags & array & Mixed \\
frequency & string & English \\
sentiment & string & English \\
historical\_significance & boolean & English \\
religious\_significance & boolean & English \\
cultural\_significance & boolean & English \\
scape & string & English \\
history & array & Mixed \\
note & string & Mixed \\
\bottomrule
\end{tabular}
\end{adjustbox}
\caption{The 19-field schema for each entry in the new idiom dataset.}
\label{tab:dataset_schema}
\end{table}

\subsection{Corpus Construction and Annotation Protocol}

The corpus was developed over five months by a dedicated, multidisciplinary team of six researchers with specific domain expertise.

\textbf{Phase 1: Data Collection and Filtering.} We aggregated an initial corpus of idioms from authoritative sources spanning three centuries, including Bangla Academy dictionaries, regional folk literature, and digital archives. The expert team then manually filtered this collection to verify authenticity and relevance, resulting in 10,361 entries that were subsequently normalized to Standard Bangla Unicode (UTF-8).

\textbf{Phase 2: AI-Assisted Annotation with Human Oversight.} We employed a human-in-the-loop workflow where AI served strictly as an assistive tool. For each entry, models like Gemini-2.5-flash and GPT-4o generated initial draft translations and sentiment suggestions. Crucially, this AI-generated content was never accepted directly. Instead, it served as a baseline for a meticulous human review process where at least two domain experts from the team would review, validate, and extensively refine or entirely replace the suggestions to ensure linguistic precision and cultural nuance.

\textbf{Phase 3: Expert Validation and Scholarly Consensus.} The final phase prioritized deep qualitative accuracy through direct engagement with a specialized expert panel, deliberately eschewing standard crowd-sourcing protocols. This panel was composed of six members, bringing a blend of academic theory and pedagogical practice: one computer scientist and five experts in Bangla language and linguistics (including academics and educators).

Validation by this group was not conducted via individual scoring rubrics; instead, consensus was actively constructed through iterative physical meetings and dedicated online discussions. This methodology was chosen to leverage the deep, non-technical expertise of the scholars, allowing for the resolution of ambiguities in a way that quantitative metrics cannot capture. Every single data point in the corpus, regardless of its origin, was subject to this review. For an idiom entry to be finalized and marked as "done" a minimum of two experts had to reach a complete agreement through these discussions. This process ensures that the final dataset reflects a deep scholarly consensus rather than a statistical aggregation of disconnected annotations.

\subsection{Experimental Evaluation: Comprehensive LLM Benchmarking}

To establish the dataset's utility as a benchmark resource, we conducted a comprehensive evaluation of 30 state-of-the-art LLMs selected through a strategic sampling methodology that ensures maximal representation across key dimensions of modern AI development.

\textbf{Model Selection Strategy.} Our model portfolio was constructed to capture the complete ecosystem of contemporary language models through deliberate stratification across four critical axes:

\begin{itemize}
    \item \textbf{Architectural Paradigms:} We included transformer variants (Llama, GPT), mixture-of-experts (Mixtral-8x22B, Mistral-Saba), and specialized architectures (Tongyi-DeepResearch, Claude-Haiku) to assess how structural differences affect cross-linguistic reasoning.
    
    \item \textbf{Scale Spectrum:} The selection spans five orders of magnitude, from compact 4.5B parameter models (Arcee-AI's AFM-4.5B) for edge deployment scenarios to massive 405B parameter models (Llama-3.1-405B) representing current scaling limits, enabling analysis of scale-performance relationships.
    
    \item \textbf{Geopolitical Distribution:} Our cohort includes models from US (OpenAI, Anthropic, Meta), Chinese (Qwen, Baidu, Alibaba), European (Mistral) and other developers, capturing diverse training data compositions and cultural biases that may impact Bangla idiom comprehension.
    
    \item \textbf{Economic and Access Models:} We balanced proprietary production systems (GPT-4.1, Gemini-2.5) with open-source alternatives (Llama series, Gemma) and research-focused models (Skyfall-36B) to evaluate how development incentives and transparency affect performance.
\end{itemize}

This strategic sampling ensures our benchmark assesses not merely individual model capabilities but reveals broader patterns about how architectural choices, scaling laws, training data diversity, and development ecosystems influence cross-cultural linguistic understanding.

\textbf{Experimental Design.} We employed a zero-shot evaluation protocol on a subset of 100 culturally significant idioms. This subset was meticulously curated through iterative expert discussions by linguistics specialists who reviewed the entire new idiom dataset, explicitly avoiding any computational or random selection. Priority was given to idioms that are not only commonly used in natural language and daily conversation but also possess subtle or complex figurative contexts, ensuring the test set represents a meaningful spectrum of linguistic difficulty. The model cohort intentionally including specialized variants (Tongyi-DeepResearch for academic contexts, MiniMax-M1 for multimodal grounding) alongside general-purpose models was identically prompted to explain each idiom's meaning and cultural connotations.

\begin{figure}[!h]
    \centering
    \includegraphics[width=1\linewidth]{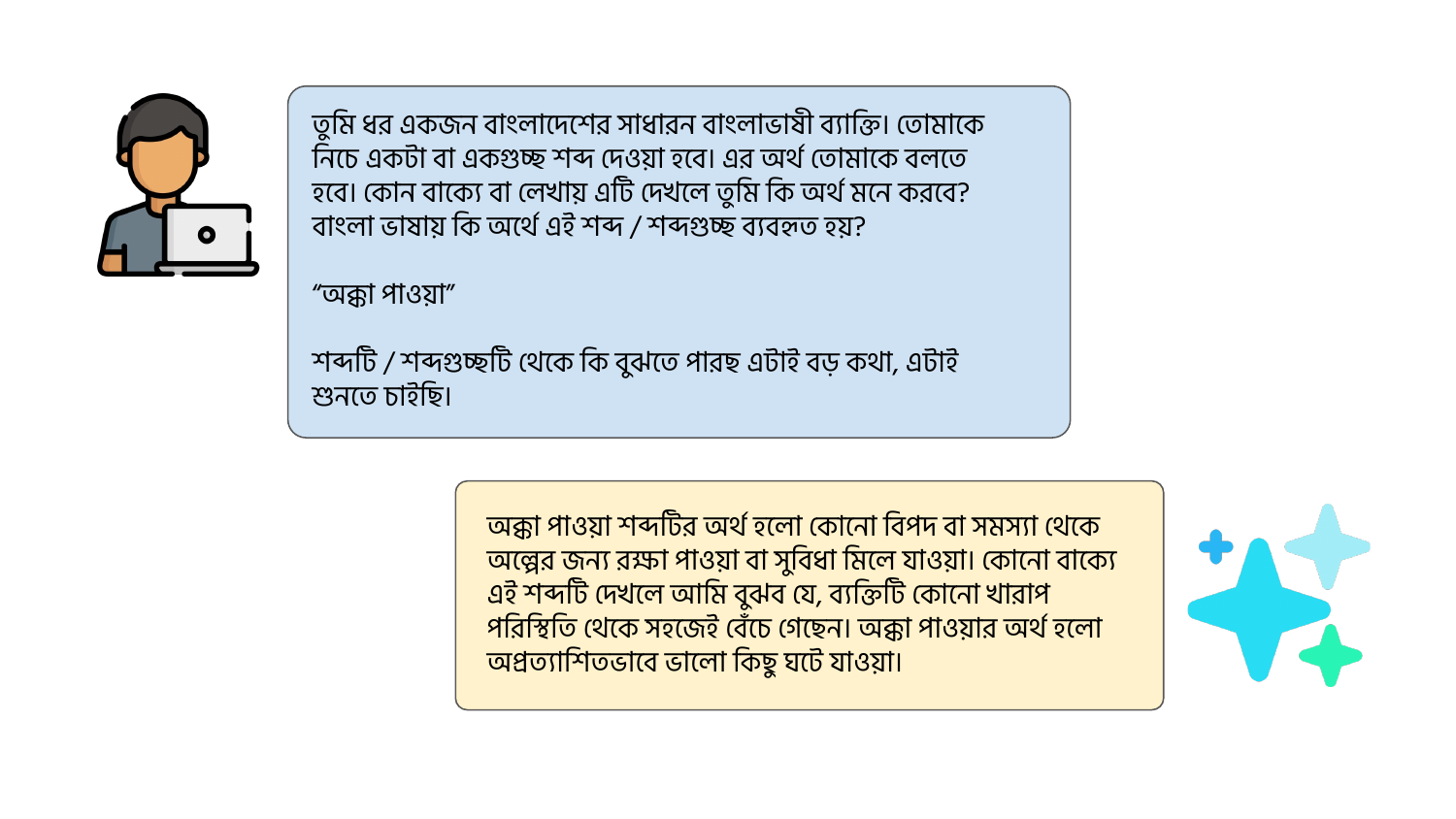}
    \caption{Example interaction with an LLM}
    \label{fig:prompt-example}
\end{figure}

\textbf{Prompt Design.} All models received identical prompts in Bangla, designed to elicit natural-language understanding from a native speaker's perspective. An example of the prompt is shown in Figure~\ref{fig:prompt-example}. As required for non-English content, the English translation of the core prompt given to the models is provided below:
\begin{quote}
``Suppose you are an ordinary Bengali-speaking person from Bangladesh. You will be given an idiom/phrase below. You have to tell its meaning. If you see it in a sentence or writing, what meaning will you think of? In what sense is this word/phrase used in the Bangla language? [...] The most important thing is what you understand from the word/phrase, that's what we want to hear.''
\end{quote}

\textbf{Evaluation Protocol.} An expert panel of five Bangla linguistics specialists conducted a double-blind review of all 3,000 responses. Each response was scored on a validated 6-point rubric assessing the semantic comprehension of the figurative meaning, detailed in Table~\ref{tab:rubric}. This rigorous protocol enables comparative analysis of model capabilities against human performance.

\begin{table}[h]
\centering
\begin{adjustbox}{max width=\columnwidth}
\begin{tabular}{cl}
\hline
\textbf{Score} & \textbf{Description of Comprehension} \\
\hline
5 & Precise and complete figurative meaning \\
4 & Correct figurative meaning \\
3 & Correct meaning with minor inaccuracies \\
2 & Vague/partial figurative grasp \\
1 & Literal meaning only \\
0 & Complete misunderstanding \\
\hline
\end{tabular}
\end{adjustbox}
\caption{Evaluation rubric for semantic comprehension of figurative meaning.}
\label{tab:rubric}
\end{table}

\subsection{Human Baseline}

To contextualize the LLM performance and establish a native-speaker benchmark, we administered the same 100-idiom question set to a cohort of 15 Bangladeshi students (6 high school and 9 undergraduate). The evaluation was conducted under controlled conditions, providing flexible time while prohibiting access to the internet or any additional textual resources.

This human cohort achieved a mean score of 417 out of 500 (83.4\%). This translates to an average score of 4.17 per question on the 6-point evaluation rubric (see Table~\ref{tab:rubric}), with a standard deviation of 0.3936. This result serves as the human baseline for figurative and cultural comprehension against which the model performances are measured.

\section{Dataset Description}

The new idiom dataset comprises \textbf{10,361 annotated Bangla idioms}, making it the largest idiomatic dataset for Bangla. This section provides key statistics and characteristics.

\subsection{Corpus Statistics and Semantic Coverage}

The dataset contains \textbf{37,129 tag assignments} with an average of \textbf{3.58 semantic tags per idiom}. Importantly, semantic tags are not drawn from a predefined classification set, but are flexibly assigned according to the nuanced semantics and contextual meanings expressed in each idiom. This approach yielded \textbf{9,919 unique semantic tags}, demonstrating exceptional semantic diversity and capturing the full spectrum of idiomatic meanings in Bangla.

\subsection{Sentiment and Domain Distribution}

Sentiment analysis reveals a \textbf{negative bias} (47.1\%, 4,882 idioms), with neutral at 37.4\% (3,877) and positive at 15.5\% (1,601), reflecting Bangla idioms' cautionary nature.

Domain coverage shows \textbf{conversation-related domains dominate} (53.1\%, 5,498 idioms), with literature at 13.6\% (1,409 idioms) and specialized domains (education, media) ensuring broad NLP applicability (Table~\ref{tab:usage_domains}).

\begin{table}[h]
\centering
\begin{tabular}{lc}
\hline
\textbf{Usage Domain} & \textbf{Percentage} \\
\hline
Everyday Conversation & 30.5\% \\
Literature & 13.6\% \\
Daily Informal Talks & 12.4\% \\
Media & 8.4\% \\
Education & 7.2\% \\
\hline
\end{tabular}
\caption{Primary Usage Domain Distribution}
\label{tab:usage_domains}
\end{table}

\subsection{Geographical and Cultural Coverage}
\begin{figure}[!h]
    \centering
    \includegraphics[width=0.9\linewidth]{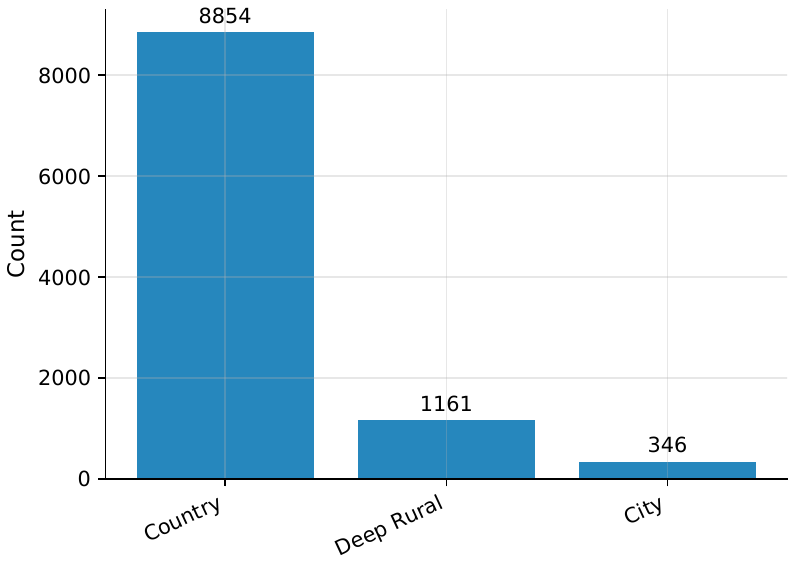}
    \caption{Dataset Geographical and Cultural Coverage}
    \label{fig:geo_cul_cov}
\end{figure}

The corpus shows extensive geographical representation: \textbf{85.5\% country-wide} (8,854 idioms), 11.2\% rural (1,161), and 3.3\% urban (346) (Figure~\ref{fig:geo_cul_cov}). Cultural annotations reveal 99.7\% have cultural significance, 13.5\% carry religious significance, and 6.3\% bear historical significance.

\subsection{Frequency and Thematic Analysis}

Frequency distribution balances accessibility and completeness: \textbf{75.1\% common idioms} (7,777) and 17.2\% rare (1,781), ensuring practical NLP applications while preserving linguistic diversity.

Thematic analysis reveals rich semantic patterns, with tags organically emerging from idiom meanings rather than being constrained by predefined categories. This flexible tagging approach captured four primary clusters: Emotional/Affective expressions, Conceptual/Literary devices, Social/Communicative functions, and Behavioral/Moral categories.

\subsection{Register and Annotation Quality}

Register distribution shows a \textbf{literary-to-conversational ratio of 1:4.6}. The dataset achieves \textbf{100\% annotation completeness} for primary schema fields and 99.7\% for cultural significance, ensuring reliable model training.

The comprehensive annotation framework, particularly the semantic tagging system that adapts to each idiom's unique meaning rather than imposing fixed categories, positions the new idiom dataset as a foundational resource for Bangla NLP in figurative language understanding and cultural computing.

\section{Model Performance Analysis}

The comprehensive evaluation of 30 state-of-the-art language models on the selected subset of the new idiom dataset benchmark reveals a striking and systemic failure in Bengali figurative language comprehension. A top-level overview of the results, detailed in Table~\ref{tab:model_results}, shows that no model surpassed the 50\% performance threshold. The leading model, \texttt{google/gemini-2.5-flash}, scored only 238 out of 500 (47.6\%), while the majority of models exhibited catastrophic failure rates, often exceeding 80-90\%. This poor performance is consistent across all model sizes, architectures, and geographic origins, pointing to a fundamental gap in current multilingual training paradigms rather than isolated model-specific weaknesses.

\subsection{Performance Stratification and Variance}

The results show a distinct stratification of model capabilities. A top tier emerges, led by \texttt{google/gemini-2.5-flash} (Mean Score: 2.38), followed by \texttt{anthropic/claude-3.5-sonnet} (1.93) and \texttt{deepseek/deepseek-chat-v3.1} (1.73). These models demonstrate a partial capacity for the task, achieving perfect scores (5/5) on 38, 30, and 23 idioms, respectively. 

However, this potential is undermined by extreme performance inconsistency. All top-tier models exhibit high standard deviations (SD > 2.15) and Coefficients of Variation (CoV) ranging from 0.99 to 1.25. Such high variance suggests that their success on certain idioms is unpredictable and not indicative of a generalizable understanding of Bengali figurative language.

\subsection{Prevalence of Task Failure}

A critical finding is the pervasive rate of task failure, defined as a score of 0, across all models. Even the leading model, \texttt{gemini-2.5-flash}, failed completely on 46 of the 100 idioms. The failure rate escalates dramatically for lower-ranked models, with over half of the evaluated models failing on more than 80\% of the test cases.

\begin{table*}[ht!]
\centering
\resizebox{\textwidth}{!}{
\begin{adjustbox}{max width=\columnwidth}
\begin{tabular}{clccccccccc}
\hline
Rank & Model & Total & Mean & Std & Min & Max & IQR & CoV & Perfect & Fail \\
\hline
1 & gemini-2.5-flash & 238 & 2.38 & 2.36 & 0 & 5 & 5.0 & 0.99 & 38 & 46 \\
2 & claude-3.5-sonnet & 193 & 1.93 & 2.25 & 0 & 5 & 5.0 & 1.17 & 30 & 53 \\
3 & deepseek-chat-v3.1 & 173 & 1.73 & 2.16 & 0 & 5 & 4.0 & 1.25 & 23 & 58 \\
4 & llama-4-maverick & 161 & 1.61 & 2.16 & 0 & 5 & 4.0 & 1.34 & 22 & 61 \\
5 & gemini-2.5-flash-lite & 145 & 1.45 & 2.18 & 0 & 5 & 4.0 & 1.50 & 24 & 67 \\
6 & chatgpt-4o-latest & 141 & 1.41 & 2.11 & 0 & 5 & 4.0 & 1.50 & 19 & 66 \\
7 & gpt-4.1 & 136 & 1.36 & 2.02 & 0 & 5 & 3.0 & 1.49 & 18 & 64 \\
8 & qwen3-max & 128 & 1.28 & 2.00 & 0 & 5 & 3.0 & 1.56 & 15 & 69 \\
9 & claude-haiku-4.5 & 109 & 1.09 & 1.85 & 0 & 5 & 2.0 & 1.70 & 10 & 72 \\
10 & llama-4-scout & 108 & 1.08 & 1.85 & 0 & 5 & 2.0 & 1.71 & 12 & 71 \\
11 & mistral-medium-3.1 & 86 & 0.86 & 1.72 & 0 & 5 & 0.0 & 2.00 & 11 & 76 \\
12 & grok-3-mini & 84 & 0.84 & 1.70 & 0 & 5 & 0.0 & 2.02 & 10 & 77 \\
13 & gemma-3n-e4b-it & 83 & 0.83 & 1.60 & 0 & 5 & 1.0 & 1.92 & 7 & 74 \\
14 & gemma-3-12b-it & 81 & 0.81 & 1.59 & 0 & 5 & 0.0 & 1.96 & 6 & 76 \\
15 & kimi-k2-0905 & 78 & 0.78 & 1.69 & 0 & 5 & 0.0 & 2.16 & 10 & 80 \\
16 & llama-3.1-405b & 68 & 0.68 & 1.58 & 0 & 5 & 0.0 & 2.33 & 8 & 83 \\
17 & grok-4-fast & 66 & 0.66 & 1.54 & 0 & 5 & 0.0 & 2.33 & 8 & 82 \\
18 & nova-pro-v1 & 49 & 0.49 & 1.34 & 0 & 5 & 0.0 & 2.74 & 4 & 87 \\
19 & tongyi-deepresearch & 48 & 0.48 & 1.28 & 0 & 5 & 0.0 & 2.67 & 5 & 84 \\
20 & minimax-m1 & 46 & 0.46 & 1.36 & 0 & 5 & 0.0 & 2.95 & 6 & 88 \\
21 & llama-3.3-70b & 36 & 0.36 & 1.02 & 0 & 5 & 0.0 & 2.83 & 3 & 85 \\
22 & glm-4.5 & 34 & 0.34 & 1.14 & 0 & 5 & 0.0 & 3.35 & 3 & 91 \\
23 & ernie-4.5-21b & 33 & 0.33 & 0.99 & 0 & 5 & 0.0 & 2.99 & 1 & 88 \\
24 & longcat-flash-chat & 30 & 0.30 & 0.90 & 0 & 4 & 0.0 & 3.02 & 0 & 88 \\
25 & mistral-saba & 22 & 0.22 & 0.81 & 0 & 5 & 0.0 & 3.69 & 1 & 92 \\
26 & cydonia-24b-v4.1 & 19 & 0.19 & 0.75 & 0 & 4 & 0.0 & 3.94 & 0 & 93 \\
27 & skyfall-36b-v2 & 4 & 0.04 & 0.24 & 0 & 2 & 0.0 & 6.07 & 0 & 97 \\
28 & mixtral-8x22b & 2 & 0.02 & 0.20 & 0 & 2 & 0.0 & 10.00 & 0 & 99 \\
29 & afm-4.5b & 0 & 0.00 & 0.00 & 0 & 0 & 0.0 & 0.00 & 0 & 100 \\
29 & qwen3-vl-8b & 0 & 0.00 & 0.00 & 0 & 0 & 0.0 & 0.00 & 0 & 100 \\
\hline
\end{tabular}
\end{adjustbox}
}
\caption{Model performance comparison across various metrics.}
\label{tab:model_results}
\end{table*}

\subsection{Analysis of Architectural and Training Patterns}

A deeper analysis of the results reveals no correlation between performance and the models' underlying characteristics, such as size, architecture, or regional origin. This suggests the problem is universal to the current generation of LLMs.

\textbf{Insignificance of Geographic Origin:} High- and low-performing models originate from developers across the US, China, and Europe. For instance, top performers include models from Google (US) and DeepSeek (China), while poor performers also come from diverse geographic backgrounds. This lack of a regional pattern indicates that current multilingual pre-training strategies, regardless of their origin, are uniformly deficient in capturing the required cultural context for Bengali.

\textbf{Impact of Model Size:} The benchmark demonstrates that larger model size does not confer a significant advantage. For example, \texttt{meta-llama/llama-3.1-405b-instruct}, one of the largest models evaluated, ranked 16th with a mean score of only 0.68. In contrast, the much smaller and more efficient \texttt{google/gemini-2.5-flash} leads the benchmark. This disparity strongly suggests that simply scaling up parameters is an insufficient strategy for improving performance on culturally-specific, low-resource language tasks.

\textbf{Architectural Paradigm:} The rankings are populated by a mix of architectural designs, including dense transformer models and Mixture-of-Experts (MoE) architectures like \texttt{mistralai/mixtral-8x22b-instruct}. No single paradigm demonstrated a clear advantage; in fact, the aforementioned MoE model was one of the worst performers, scoring only 0.02. This implies that the performance deficit is not an architectural flaw but a data and training issue. Furthermore, specialized models like the research-oriented \texttt{alibaba/tongyi-deepresearch-30b} (rank 19) and the multimodal \texttt{qwen/qwen3-vl-8b-instruct} (rank 29) also failed to perform, confirming that neither academic specialization nor visual grounding currently compensates for this core linguistic gap.

\section{Limitations and Future Directions}

The development of the new idiom dataset provides a robust foundation for Bangla computational linguistics and illuminates several key areas for future enhancement. The current resource's limitations suggest a clear roadmap for advancing the field, where further work can be done in the following directions:

\begin{itemize}
    \item \textbf{Temporal and Generational Expansion:} Future work can capture contemporary digital neologisms for diachronic study and develop generation-aware frameworks to analyze how idiom interpretation varies across demographics.

    \item \textbf{Geographical and Dialectical Representation:} Coverage can be deepened to include underrepresented dialects (e.g., Sylheti, Chittagonian) and urban linguistic innovations, facilitating robust dialect-aware systems.

    \item \textbf{Annotation Frameworks and Scalability:} The 19-field schema can be expanded using semi-supervised or community-sourced methods for efficient scaling, alongside developing more nuanced metrics for cultural significance.

    \item \textbf{Sentiment Analysis and Domain Adaptation:} The observed sentiment distribution (47.1\% negative) offers a distinct opportunity to develop robust domain adaptation techniques for sentiment analysis models.

    \item \textbf{Benchmark Development:} A significant opportunity exists to establish standardized benchmarks and evaluation tasks (detection, interpretation, generation) to systematize research in Bangla idiom processing.

    \item \textbf{Benchmark Scope and Evaluation Robustness:} The 100-idiom test set can be significantly expanded to ensure robustness, while objective metrics can be investigated to complement subjective, consensus-based expert evaluations.

    \item \textbf{Broadening Model Diversity in Benchmarking:} The benchmark can be continually updated with a wider array of models, new architectures, and training paradigms beyond the current 30 to better track state-of-the-art capabilities.

    \item \textbf{Computational Efficiency and Accessibility:} Lightweight access methods, specialized data subsets, and APIs can be developed to support researchers in resource-constrained environments.

    \item \textbf{Annotation Rigor:} Future annotation efforts provide an opportunity to compute formal agreement metrics (e.g., Cohen's $\kappa$) through parallel annotation, complementing the current qualitative consensus method.
\end{itemize}

In summary, each identified limitation of the new idiom dataset presents a clear opportunity not only to enhance Bangla figurative language resources but also to advance NLP methodologies for low-resource languages globally.

\section{Conclusion}

As the first large-scale corpus of its kind, the new idiom dataset contains 10,361 idioms annotated with a detailed 19-field schema designed to capture their rich semantic, pragmatic, and cultural nuances. This resource addresses a long-standing gap in figurative language analysis for low-resource languages.

Our benchmark evaluations reveal a significant performance gap in state-of-the-art large language models, none of which surpassed 50\% accuracy on Bangla idiom comprehension. This finding highlights the urgent need for language models that are more deeply attuned to cultural and figurative contexts.

The development of the new idiom dataset is an ongoing initiative. We are committed to continuously expanding and refining the corpus to enhance its utility for diverse NLP applications, such as machine translation, cultural analytics, and educational technology. By making this resource publicly available, we aim to stimulate innovation in Bangla NLP and encourage the creation of similar culturally-grounded resources for other underrepresented languages, contributing to a more inclusive and linguistically aware AI ecosystem.


\bibliographystyle{lrec2026-natbib}
\bibliography{bibliography}


\end{document}